\documentclass[a4paper,fleqn]{cas-dc}

\usepackage[]{natbib}

\def\tsc#1{\csdef{#1}{\textsc{\lowercase{#1}}\xspace}}
\tsc{WGM}
\tsc{QE}

\usepackage{amssymb}
\usepackage{hyperref}       

\usepackage{subcaption}

\usepackage{algorithm}
\usepackage{array}
\usepackage{url}            
\usepackage{booktabs}       
\usepackage{verbatim}
\usepackage{stfloats}
\usepackage{graphicx}
\usepackage{amsmath,amsfonts,amssymb,amsthm}

\usepackage{enumitem}
\usepackage{xcolor}
\usepackage{hyperref}
\usepackage{url}
\usepackage{booktabs}
\usepackage{graphicx}
\usepackage{eurosym}
\usepackage{multirow}
\usepackage{dsfont}
\usepackage{rotating}
\usepackage{tabularray}
\usepackage{algpseudocode}

\hypersetup{
    colorlinks = true,
    citecolor = blue,
    linkcolor = blue,
    urlcolor = blue,
    bookmarksopen = true,
    filecolor=blue
}

\usepackage{pifont}
\newcommand{\cmark}{\ding{51}}%
\newcommand{\xmark}{\ding{55}}%

\newcommand{\soc}{\text{SoC}}

\newcommand{\ch}{\text{ch}}
\newcommand{\dis}{\text{dis}}
\newcommand{\cs}{\text{cs}}

\begin{document}
\let\WriteBookmarks\relax
\def\floatpagepagefraction{1}
\def\textpagefraction{.001}

\shorttitle{GNN-DT: A Graph Neural Network Enhanced Decision Transformer for Efficient Optimization in Dynamic Environments}    

\shortauthors{Orfanoudakis et al.}  

\title [mode = title]{GNN-DT: A Graph Neural Network Enhanced Decision Transformer for Efficient Optimization in Dynamic Environments} 

\author[]{Stavros~Orfanoudakis}[orcid=0000-0002-0767-9488]
\cormark[1]
\cortext[1]{Corresponding author email: s.orfanoudakis@tudelft.nl}
\credit{Conceptualization, Methodology, Software, Writing - Original Draft, Writing - Review \& Editing}

\affiliation[]{organization={Delft University of Technology, Intelligent Electrical Power Grids},
            addressline={Mekelweg 5},
            city={Delft},
            postcode={2628 CD},            
            country={The Netherlands}}

\author[]{Nanda~Kishor~Panda}[orcid=0000-0002-9647-4424]
\credit{Conceptualization, Writing - Original Draft, Writing - Review \& Editing}
\author[]{Peter~Palensky}[orcid=0000-0003-3183-4705]
\credit{Supervision, Funding acquisition}
\author[]{Pedro~P.~Vergara}[orcid=0000-0003-0852-0169]
\credit{Conceptualization, Supervision, Funding acquisition, Writing - Review \& Editing}

\begin{abstract}
Reinforcement Learning (RL) methods used for solving real-world optimization problems often involve dynamic state-action spaces, larger scale, and sparse rewards, leading to significant challenges in convergence, scalability, and efficient exploration of the solution space. This study introduces GNN-DT, a novel Decision Transformer (DT) architecture that integrates Graph Neural Network (GNN) embedders with a novel residual connection between input and output tokens crucial for handling dynamic environments. By learning from previously collected trajectories, GNN-DT tackles the sparse rewards limitations of online RL algorithms and delivers high-quality solutions in real-time. We evaluate GNN-DT on the complex electric vehicle (EV) charging optimization problem and prove that its performance is superior and requires significantly fewer training trajectories, thus improving sample efficiency compared to existing DT and offline RL baselines. Furthermore, GNN-DT exhibits robust generalization to unseen environments and larger action spaces, addressing a critical gap in prior offline and online RL approaches.
\end{abstract}




\maketitle

\section{Introduction}

Sequential decision-making problems are critical for efficiently operating a wide array of industries, such as power systems control~\cite{ROALD2023108725}, logistics optimization~\cite{konstantakopoulos_vehicle_2022}, portfolio management~\cite{gunjan2023brief}, and advanced manufacturing processes~\cite{2020OptimizationOM}. However, many practical problems, such as the electric vehicle (EV) charging optimization~\cite{n2024quantifying}, are large-scale, have temporal dependencies, and aggregated constraints, often making conventional methods impractical~\cite{8187196}. 
This is especially observed in dynamic environments, where the optimization landscape continuously evolves, requiring real-time solutions.

Reinforcement learning (RL)~\cite{SuttonReinforcementIntroduction} has been extensively studied for solving optimization problems due to its ability to manage uncertainty, adapt to dynamic environments, and enhance decision-making through trial-and-error~\cite{lan2023learning, ZHANG2023205}. In complex and large-scale scenarios, RL can provide high-quality solutions in real-time compared to traditional mathematical programming techniques that fail to do so~\cite{jaimungal_reinforcement_2022}. However, RL approaches face significant challenges, such as sparse reward signals that slow learning and hinder convergence to optimal policies~\cite{dulac-arnold_challenges_2021}. In addition, RL solutions struggle to generalize when deployed in environments different from the one they were trained in, limiting their applicability in real-world scenarios with constantly changing conditions~\cite{wang_domain_2024}.

Decision Transformers (DT)~\cite{chen2021decision} is an offline RL algorithm that reframes traditional RL problems as generative sequence modeling tasks conditioned on future rewards~\cite{Zhang2023}.
DTs utilize previously collected trajectories to inform decision-making without requiring an environment simulator.
By learning from historical data, DTs effectively address the sparse reward issue inherent in online RL, relying on demonstrated successful outcomes instead of extensive trial-and-error exploration. However, the trajectory-stitching mechanism of DT often proves insufficient in dynamic real-world environments, leading to suboptimal policies. Although improved variants such as Q-regularized DT (Q-DT)~\cite{qdt} incorporate additional constraints for greater robustness, they still face significant challenges in generalizing across non-stationary tasks~\cite{10.5555/3600270.3603094}. Consequently, further architectural advances and training strategies are essential to ensure consistent performance in complex environments.

This study introduces GNN-DT\footnote{The code can be found at \url{https://github.com/StavrosOrf/DT4EVs}.}, a novel DT architecture that leverages the permutation-equivariant properties of Graph Neural Networks (GNNs) to handle dynamically changing state-action spaces (i.e., varying numbers of nodes over time) and improve generalization. By generating embeddings that remain consistent under node reordering, GNNs offer a powerful way to capture relational information in complex dynamic environments. Moreover, {GNN-DT} features a novel residual connection between input and output tokens, ensuring that action outputs are informed by the dynamically learned state embeddings for more robust decision-making. 
To demonstrate the superior performance of the proposed method, we conduct extensive experiments on the complex multi-objective EV charging optimization problem~\cite{10803908}, which encompasses sparse rewards, temporal dependencies, and aggregated constraints.
The main contributions are summarized as follows:
\begin{itemize}
    \item Introducing a novel DT architecture integrating GNN embeddings, resulting in enhanced sample efficiency, superior performance, robust generalization to unseen environments, and effective scalability to larger action spaces, demonstrating the critical role of GNNs.
    \item Demonstrating that online and offline RL baselines, even when trained on diverse datasets (Optimal, Random, Business-as-Usual) with varying sample sizes, perform inferior to GNN-DT when dealing with real-world optimization tasks.
    \item Proving that both the size and type of training dataset critically influence the learning process of DTs, highlighting the importance of dataset selection. Also, shown that strategically integrating high- and low-quality training data (Optimal \& Random datasets) significantly enhances policy learning, outperforming models trained exclusively on single-policy datasets.
\end{itemize}

\section{Related Work}

\subsection{Advancements in Decision Transformers}
Classic DT encounters significant challenges, including limited trajectory stitching capabilities and difficulties in adapting to online environments. To address these issues, several enhancements have been proposed. The Q-DT~\cite{qdt} improves the ability to derive optimal policies from sub-optimal trajectories by relabeling return-to-go values in the training data. Elastic DT~\cite{10.5555/3666122.3666936} enhances classic DT by enabling trajectory stitching during action inference at test time, while Multi-Game DT~\cite{10.5555/3600270.3602295} advances its task generalization capabilities. The Online DT~\cite{online_dt,3056976d4116471a86ab3fa345b1695d} extends DTs to online settings by combining offline pretraining with online fine-tuning, facilitating continuous policy updates in dynamic environments. Additionally, adaptations for offline safe RL incorporate cost tokens alongside rewards~\cite{liu2023constrained, pmlr-v238-hong24a}. DT has also been effectively applied to real-world domains, such as healthcare~\cite{Zhang2023} and chip design~\cite{pmlr-v202-lai23c}, showcasing its practical utility.

\subsection{RL for EV Smart Charging}
RL algorithms offer notable advantages for EV dispatch, including the ability to handle nonlinear models, robustly quantify uncertainty, and deliver faster computation than traditional mathematical programming~\cite{QIU2023113052}. Popular methods, such as Deep Deterministic Policy Gradient (DDPG)~\cite{JIN2022120140}, Soft Actor Critic (SAC)~\cite{9211734}, and batch RL~\cite{8727484}, show promise but often lack formal constraint satisfaction guarantees and struggle to scale with high-dimensional state-action spaces~\cite{isgt2024, 9465776}. Safe RL frameworks address these drawbacks by imposing constraints via constrained MDPs, but typically sacrifice performance and scalability~\cite{ZHANG2023121490, chen2022deep}. Multiagent RL techniques distribute complexity across multiple agents, e.g., charging points, stations, or aggregators~\cite{KAMRANI2025100620}, yet still encounter convergence challenges and may underperform in large-scale applications. 
To the best of our knowledge, no study has used DTs to solve the complex EV charging problem, despite DTs' potential to handle sparse rewards effectively.

\section{Problem Formulation}

\begin{table*}[!t]
\centering
\small
\caption{Notation for the EV Charging Optimization Problem}
\begin{tabular}{lll}
\toprule
\textbf{Symbol} & \textbf{Name} & \textbf{Description} \\
\midrule
\multicolumn{3}{l}{\textit{Sets}} \\
$\mathcal{T}$ & Set of timesteps & Time horizon for optimization \\
$\mathcal{I}$ & Set of charging stations & All EV charging stations \\
$\mathcal{W}$ & Set of charger groups & Chargers grouped by local transformer connections \\
$\mathcal{J}_i$ & Set of charging sessions & Charging sessions at charger $i$ \\
\midrule
\multicolumn{3}{l}{\textit{Indexes}} \\
$t$ & Timestep index & Discrete time step \\
$i$ & Charger index & Individual EV charging station \\
$j$ & Session index & Charging session at a charger \\
$w$ & Charger group index & Charger groups connected to transformers \\
\midrule
\multicolumn{3}{l}{\textit{Parameters}} \\
$t^a_{j,i}$ & Arrival time & Time EV $j$ arrives at charger $i$ \\
$t^d_{j,i}$ & Departure time & Time EV $j$ departs charger $i$ \\
$e^*_{j,i}$ & Desired battery capacity & Desired battery energy at departure for session $j$ at charger $i$ \\
$e^a_{j,i}$ & Arrival battery energy & Battery energy at EV arrival \\
$\underline{e}_{j,i}, \overline{e}_{j,i}$ & Battery limits & Min/max allowable battery energy \\
$\underline{p}^{+}_{j,i}, \overline{p}^{+}_{j,i}$ & Charging power limits & Min/max charging power \\
$\underline{p}^{-}_{j,i}, \overline{p}^{-}_{j,i}$ & Discharging power limits & Min/max discharging power \\
$p^*_t$ & Total power limit & Desired aggregated power \\
$\Pi^{+}_t, \Pi^{-}_t$ & Electricity prices & Prices for charging/discharging at timestep $t$ \\
$\Delta t$ & Time interval & Duration of each timestep \\
$\overline{p}_{w,t}$ & Group power limit & Power limit for group $w$ at timestep $t$ \\
\midrule
\multicolumn{3}{l}{\textit{Variables}} \\
$p^+_{i,t}, p^-_{i,t}$ & Charging/discharging power & Power assigned at charger $i$, timestep $t$ \\
$\omega^+_{i,t}, \omega^-_{i,t}$ & Binary operation indicators & Indicates if charger $i$ charges ($+$) or discharges ($-$) at timestep $t$ \\
$e_{j,i,t}$ & EV battery energy & Battery level for session $j$ at charger $i$, timestep $t$ \\
$p^{\sum}_t$ & Total aggregated power & Net total power across all chargers at timestep $t$ \\
\bottomrule
\end{tabular}
\label{tab:notation}
\end{table*}

In this section, an introduction to offline RL and the mathematical formulation of the EV charging optimization problem is presented as an example of what type of problems can be solved by the proposed GNN-DT methodology.

\subsection{Offline RL}

Offline RL aims to learn a policy \(\pi_\theta(a\!\mid\!s)\) that maximizes the expected discounted return \(\mathbb{E}\bigl[\sum_{t=0}^{\infty} \gamma^t R(s_t,a_t)\bigr]\)
without additional interactions with the environment~\cite{levine2020offline}. 
A Markov Decision Process (MDP) is defined by the tuple 
\((S, A, P, R, \gamma)\), 
where \(S\) is the state space, \(A\) the action space, \(P\) the transition function, 
\(R\) the reward function, \(\gamma \in (0,1]\) the discount factor~\cite{SuttonReinforcementIntroduction}.
In the offline setting, a static dataset 
\(\mathcal{D} = \{(s, a, r)\}\),
collected by a (potentially suboptimal) policy, is provided.
DTs leverage this dataset by treating
RL trajectories as sequences, learning to predict actions that
maximize returns based on previously collected experiences.
A key component in DTs is the 
\emph{return-to-go} (RTG), which for a time step \(t\) can be defined as:
    $G_t = \sum_{\tau=t}^{T} \gamma^{\tau-t} \, r_\tau,$
representing the discounted cumulative reward from \(t\) 
until the terminal time \(T\). Offline RL is particularly 
beneficial when real-time exploration is costly or impractical, 
while sufficient historical data are available.

\subsection{The EV Smart Charging Problem}
\label{sec:mip}
Working closely with a charge point operator (CPO), it was evident that existing heuristic and mathematical programming charging strategies don’t scale efficiently as EV fleets grow. To address this, we designed the state–action space and objectives around real‐world operational constraints and assumptions provided by the CPO.
We consider a set of $\mathcal{I}$  charging stations indexed i, all assumed to be controlled by a CPO over a time window $\mathcal{T}$, divided into $T$ non-overlapping intervals.
Since the chargers can be spread around the city, there are charger groups $w \in \mathcal{W}$, that can have a lower-level aggregated power limits representing connections to local power transformers.
For a given time window, each charging station $i$ operates a set of  $\mathcal{J}$ non-overlapping charging sessions, denoted by $\mathcal{J}_i=\{j_{1,i}, \cdots, j_{J_i,i}\}$, where $j_{j,i}$ represents the $j^{th}$ charging event at the $i^{th}$ charging station and $J_i=|\mathcal{J}_i|$ is the total number of charging sessions seen by charging station i in an episode. A charging session is then represented as $j_{j,i}: \{t^a_{j,i},t^d_{j,i},\bar{p}_{j,i}, e^*_{j,i}\}, ~ \forall j,i$, where
$t^a, t^d, \bar{p}$ and $e^*$ represent the arrival time, departure time, maximum charging power, and the desired battery energy level at the departure time. The primary goal is to minimize the total energy cost given by:
\begin{equation}
f_1(p^+,p^-)
= \sum\nolimits_{t\in\mathcal T}\sum\nolimits_{i\in\mathcal I}
\Delta t\bigl(\Pi^+_t\,p_{i,t}^+ - \Pi^-_t\,p_{i,t}^-\bigr)
\label{eq:mo1}
\end{equation}
$p^+_{i,t}$ and $p^-_{i,t}$ denote the charging or discharging power of the $i^{th}$ charging station during time interval $t$.
$\Pi^+_t$ and $\Pi^-_t$  are the charging and discharging costs, respectively. Along with minimizing the total energy costs 
, the CPO also wants the aggregate power of all the charging stations ($p^{\sum}_t=\sum_{i\in \mathcal{I}}p^+_{i,t}-p^-_{i,t}$) to remain below the set power limit $p_t^*$. By doing so, the CPO avoids paying penalties due to overuse of network capacity. 
Hence, we introduce the penalty:
\begin{equation}
f_2(p^+,p^-)
= \sum\nolimits_{t\in\mathcal T}\max\{0,\,p_t^{\sum}-p^*_t\},
\label{eq:mo2}
\end{equation}
Maintaining the desired battery charge at departure is crucial for EV user satisfaction. We model this behavior as:
\begin{equation}
f_3(p^+,p^-)
= \sum\nolimits_{i\in\mathcal I}\,\sum\nolimits_{j\in\mathcal J_i}
\Bigl(
  \sum\nolimits_{t=t^a_{j,i}}^{\,t^d_{j,i}}(p_{i,t}^+ - p_{i,t}^-)
  - e^*_{j,i}
\Bigr)^{2}
\label{eq:mo3}
\end{equation}
Eq.~\eqref{eq:mo3} defines a sparse reward added at each EV departure based on its departure energy level. Building on the objective functions described by Eqs. \eqref{eq:mo1}-\eqref{eq:mo3}, the overall EV charging problem is formulated as a mixed integer programming (MIP) problem, subject to lower-level operational constraints (e.g., EV battery, power levels) as detailed below:
\begin{equation}
\label{eq:mo4}
\begin{aligned}
\max_{p^+,\omega^+,p^-,\omega^-}\ \sum_{t\in\mathcal{T}}\Big[&
  -100\,\max\{0,\ p_t^{\sum}-p_t^*\} \\
&\hspace{-5mm} + \sum_{i\in\mathcal{I}}\Big(
    \Delta t\big(\Pi_t^{+} p_{i,t}^{+}\omega_{i,t}^{+}-\Pi_t^{-} p_{i,t}^{-}\omega_{i,t}^{-}\big) \\
& 
\hspace{-16mm}
   -10\!\sum_{j\in\mathcal{J}_i}\!\Big(
      \sum_{\tau=t_{j,i}^a}^{t_{j,i}^d}\!\big(p_{i,\tau}^{+}\omega_{i,\tau}^{+}-p_{i,\tau}^{-}\omega_{i,\tau}^{-}\big)
      - e_{j,i}^*
   \Big)^{\!2}\Big)\Big]
\end{aligned}
\end{equation}

Subject to:
\begin{flalign}
&\overline{p}_{w,t} \geq \sum_{i\in \mathcal{W}_i}p^+_{i,t} \cdot \omega^+_{i,t} -p^-_{i,t} \cdot \omega^-_{i,t}
  & \forall  i ,\;\forall w, \;\forall t &
    \label{eq:opt3.01}
\end{flalign}
\begin{flalign}
& \underline{e}_{j,i} \leq e_{j,i,t} \leq \overline{e}_{j,i} & \forall  j, \;\forall  i , \;\forall t&
    \label{eq:opt3.3} 
\end{flalign}
\begin{flalign}
& e_{j,i,t} = e_{j,i,t-1} + (p^+_{i,t} \cdot \omega^+_{i,t} + p^-_{i,t} \cdot \omega^-_{i,t}) \cdot \Delta t &\hspace{-1mm}\forall  j, \;\forall  i , \;\forall t&
\label{eq:opt3.4}
\end{flalign}
\begin{flalign}
& e_{j,i,t} = e^{a}_{j,i} &  \forall  j, \;\forall  i , \;\forall t | \; t = {t}^a_{j,i}&
\label{eq:opt3.5}
\end{flalign}
\begin{flalign}
&\underline{p}^{+}_{j,i}\leq p^+_{i,t} \leq 
 \overline{p}^{+}_{j,i} &  \forall  j, \;\forall  i , \;\forall t&
\label{eq:opt3.6}
\end{flalign}
\begin{flalign}
&\underline{p}^{-}_{j,i} \geq p^-_{i,t} \geq 
 \overline{p}^{-}_{j,i} &  \forall  j, \;\forall  i , \;\forall t&
\label{eq:opt3.7}
\end{flalign}
\begin{flalign}
& \omega^+_{i,t} + \omega^-_{i,t} \leq 1 &  \forall  i , \;\forall t&
 \label{eq:opt3.11}
 \end{flalign}
The multi-objective optimization function in Eq.\eqref{eq:mo4} integrates Eqs. \eqref{eq:mo1}–\eqref{eq:mo3} using experimentally determined coefficients based on practical importance. 
The power of a single charger $i$ is modeled using four decision variables, $p^+ \cdot \omega^+$ and $p^- \cdot \omega^-$, where $\omega^+$ and $\omega^-$ are binary variables, to differentiate between charging and discharging behaviors and enable charging power to get values in ranges $0\cup [\underline{p}^+,\overline{p}^+]$, and discharging power in $[\underline{p}^-,\overline{p}^-]\cup0$.
Eq.~\eqref{eq:opt3.01} defines the locally aggregated transformer power limits $\overline{p}$ for chargers belonging to groups $\mathcal{W}_i$.
Eqs.~\eqref{eq:opt3.3}-\eqref{eq:opt3.5} address EV battery constraints during operation with a minimum and maximum capacity of $\underline{e}$, $\overline{e}$, and energy $e^{a}$ at time of arrival $t^{a}$. Equations \eqref{eq:opt3.6} and \eqref{eq:opt3.7} impose charging and discharging power limits for every charger-EV session combination.
To prevent simultaneous charging and discharging, the binary variables $\omega^\ch$ and $\omega^\dis$ are constrained by \eqref{eq:opt3.11}.

\subsection{EV Charging MDP}

The optimal EV charging problem can be framed as an MDP: $\mathcal{M} = (\mathcal{S}, \mathcal{A}, \mathcal{P}, R)$,
where $S$ is the state space, $A$ is the action space, $P$ is the transition probability function, and $R$ is the reward function.
At any time step $t$, the state $\boldsymbol{s}_t \in \mathcal{S}$ is represented by a dynamic graph $\mathcal{G}_t = (\mathcal{N}_t, \mathcal{E}_t)$, where $\mathcal{N}_t$ is the set of nodes and $\mathcal{E}_t$ is the set of edges. 
The graph is dynamic since the number of nodes in the state and action graph can vary in each step, because of EVs' arrival and departures.
Each node $n \in \mathcal{N}_t$ has a feature vector $\boldsymbol{x}_{n,t} \in \mathbb{R}^d$, capturing node-dependent information such as power limits and prices.
This graph structure~\cite{Orfanoudakis2024} efficiently models evolving relationships among EVs, chargers, and the grid infrastructure.
In detail, an EV node’s feature vector:
\begin{equation}
\boldsymbol{x}^{\mathrm{ev}}_t = \bigl[\soc_t,\;t^{\rm d}-t,\;j,\;i,\;w\bigr]
\end{equation}
encodes its current state of charge \(\soc_t\in[0,1]\), the remaining steps until departure \(t^{\rm d}-t\), and the identifiers \(j\), \(i\), and \(w\) of the charging point, charger, and transformer group to which it is connected.  Exact SoC and departure‐time information is assumed to be available via ISO 15118 V2G signaling.
A charging station (cs) node \(i\) is described by:
\begin{equation}
\boldsymbol{x}^{\mathrm{cs}}_t = \bigl[\overline p^+_{i,j},\;\overline p^-_{i,j},\;i\bigr],
\end{equation}
where \(\overline p^+\) and \(\overline p^-\) are its maximum charging and discharging power and \(i\) its unique identifier. 
A transformer (tr) node \(w\) has a feature vector:
\begin{equation}
\boldsymbol{x}^{\mathrm{tr}}_t = \bigl[\overline p_{w,t},\;w\bigr],
\end{equation}
where \(\overline p_{w,t}\) denotes its maximum power capacity and \(w\) its identifier, ensuring the policy respects feeder‐level constraints.
The CPO (system) node aggregates temporal and economic context in  
\begin{equation}
\boldsymbol{x}^{\mathrm{cpo}}_t = \Bigl[\tfrac{d}{7},\;\sin\!\bigl(\tfrac{h}{48\pi}\bigr),\;\cos\!\bigl(\tfrac{h}{48\pi}\bigr),\;\Pi^+_t,\;p^{\sum}_{t-1}\Bigr].
\end{equation} 
Here \(d\) and \(h\) encode day‐of‐week and hour to capture demand patterns, \(\Pi^+_t\) is the charging price at \(t\), and \(p^{\sum}_{t-1}\) is the previous time step’s total power draw, providing feedback for profit‐maximizing decisions.

The action space $\mathbf{a}_t \in \mathcal{A}$ is represented by a dynamic graph $\mathcal{G}^{\mathbf{a}}_t=(\mathcal{N}^{\mathbf{a}}_t,\mathcal{E}^{\mathbf{a}}_t)$, where nodes $\mathcal{N}^{\mathbf{a}}_t$ correspond to the decision variables of the optimization problem (e.g., EVs).  Each node $n \in \mathcal{N}^{\mathbf{a}}_t$ represents a single action $a_{i,t} \in \mathbf{a}_t$, scaled by the corresponding charger's maximum power limit.  For charging, $a_{i,t} \in [0,1]$, and for discharging, $a_{i,t} \in [-1,0)$. The transition function $\mathcal{P}(\boldsymbol{s}_{t+1} \mid \boldsymbol{s}_t, \mathbf{a}_t)$ accounts for uncertainties in EV arrivals, departures, energy demands, and grid fluctuations. 
Finally, the reward function is the same with the objective described in Eq.~\eqref{eq:mo4} and is defined as $R(\boldsymbol{s}_t, \mathbf{a}_t)= f_1 - 100 f_2 - 10 f_3$ for a single timestep ($\mathcal{T} = \{t\}$).
The reward is guiding the policy to maximize cost savings, respect operational constraints, and meet EV driver requirements. While Eqs.\eqref{eq:mo1} and \eqref{eq:mo2} represent individual EV rewards and aggregated EV penalties, respectively. Eq.\ref{eq:mo3} introduces a sparse reward that activates only when an EV departs, thereby creating complex temporal dependencies.
The reward function weights were selected after comprehensive experimentation, ensuring the optimal solution of the MIP remains the same while the RL training is balanced between exploration and penalization.

\section{GNN-based Decision Transformer}

\begin{figure*}[!t]
\centering
\includegraphics[width=\linewidth]{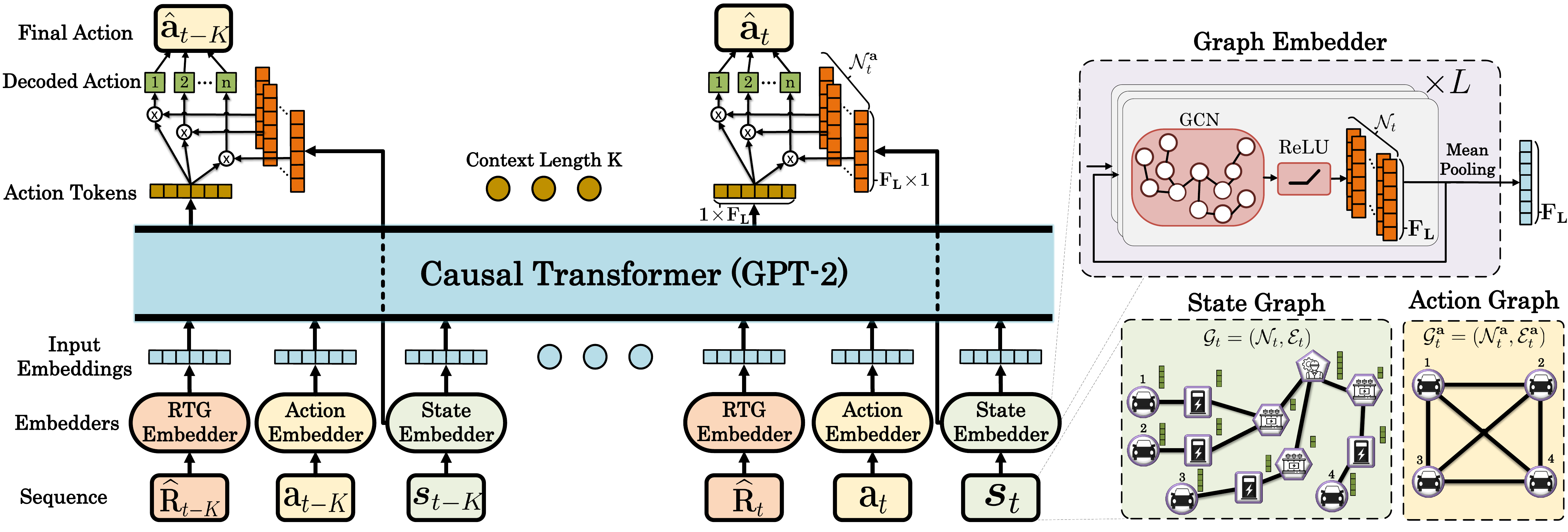}
\caption{Overview of the GNN-DT architecture. The input sequence, comprising return-to-go, action, and state, is processed through specialized embedding modules. The action graph $\mathcal{G}^{\mathbf{a}}_t=(\mathcal{N}^{\mathbf{a}}_t,\mathcal{E}^{\mathbf{a}}_t )$, with nodes $\mathcal{N}^{\mathbf{a}}_t \subset \mathcal{N}_t$, and the state graph $\mathcal{G}_t=(\mathcal{N}_t,\mathcal{E}_t )$ are encoded using GNN-based embedders to produce embeddings of dimension $F_L$. These embeddings serve as inputs to a GPT-2–based causal transformer, which predicts the next action token. The predicted action token acts as a decoder, generating actions by multiplying with specific GNN state node embeddings. 
}
\label{fig:gnndt_ow}
\end{figure*}



The innovative GNN-DT architecture (Fig.~\ref{fig:gnndt_ow}) efficiently solves optimization problems in complex environments with dynamic state-action spaces by embedding past actions, states, and returns-to-go, using a causal transformer to generate action tokens, and integrating these with current state embeddings to determine final actions within the dynamically changing action space.

\subsection{Sequence Embeddings}

In GNN-DT, each input ``modality'' is processed by a specialized embedding network. The state graph passes through the \emph{State Embedder}, the action through the \emph{Action Embedder}, and the return-to-go value through a simple Multi-Layer Perceptron (MLP). Compared to standard MLP embedders, GNNs provide embeddings for states and actions invariant to the number of nodes by capturing the graph structure. This design makes GNN-DT more sample-efficient during training and better at generalizing to unseen environments.

In detail, the \textit{State Embedder} consists of $L$ consecutive Graph Convolutional Network (GCN)~\cite{kipf2016semisupervised} layers, which aggregate information from neighboring nodes as follows:
\begin{equation}
    \boldsymbol{x}_t^{(l+1)} = \sigma \Bigl(D^{-1/2} A_t D^{-1/2} \boldsymbol{x}_t^{(l)} W^{(l)}\Bigr),
\end{equation}
where $\boldsymbol{x}_t^{(l)} \in \mathbb{R}^{N_t \times F_l}$ denotes the node embeddings at layer $l$ with $N_t$ number of nodes, $W^{(l)} \in \mathbb{R}^{F_l \times F_{l+1}}$ are trainable weights, $\sigma(\cdot)$ is a nonlinear activation (ReLU), $A_t$ is the adjacency matrix of the state graph $\mathcal{G}_t$, and $D$ is the degree matrix for normalization. After the final layer, a mean-pooling operation produces a fixed-size state embedding:
    $\boldsymbol{\widetilde{s}}_{t} = \frac{1}{\lvert \mathcal{N}_t \rvert} \sum_{n \in \mathcal{N}_t} \boldsymbol{x}_{n,t}^{(L)},$
where $\boldsymbol{x}_n^{(L)}$ is the embedding of node $n$ at the $L$-th layer. This pooling step ensures that the state embedding is invariant to the number of nodes in the graph, enabling the architecture to scale with any number of EVs or chargers.
Similarly, the \textit{Action Embedder} processes the action graph $\mathcal{G}^{\mathbf{a}}_t=(\mathcal{N}^{\mathbf{a}}_t,\mathcal{E}^{\mathbf{a}}_t)$ through $C$ GCN layers followed by mean pooling, producing the action embedding $\widetilde{\textbf{a}}_{t}$. All embedding vectors (states, actions, or the return-to-go value) have the same dimensions. 
This design leverages the dynamic and invariant nature of GCN-based embeddings, allowing the DT to handle variable-sized graphs.

\subsection{Decoding Actions}

Once the embedding sequence of length $K$ is constructed\footnote{During inference the action ($\mathbf{a}_t$) and RTG ($\boldsymbol{\widehat{R}}_t$) of the last step $t$ are filled with zeros as they are not known.}, it is passed through the causal transformer GPT-2 to produce a fixed-size output vector 
$\boldsymbol{y}_t \in \mathbb{R}^{F_L}$ for each step. Because DT architectures inherently generate outputs of fixed dimensions, an additional mechanism is required to manage dynamic action spaces. To address this, GNN-DT implements a residual connection that merges the final GCN layer embeddings $\boldsymbol{x}_t^{(L)}$ with the transformer output $\boldsymbol{y}_t$ for every step of the sequence.
Specifically, 
for each node $n \in \mathcal{N}_t^{\mathbf{a}}$, we retrieve its corresponding state embedding $\boldsymbol{x}_{n,t}^{(L)} \in \mathbb{R}^{1\times F_L}$ and multiply it with the transformer output token $\boldsymbol{y}_t\in \mathbb{R}^{1\times F_L}$, yielding the final action for node $n$:
    $\hat{\mathrm{a}}_{n,t} =  \boldsymbol{y}_t ^\mathsf{T} \cdot \boldsymbol{x}_{n,t}^{(L)}.$
By repeating for every step $t$ and every node $n \in \mathcal{N}^{\mathbf{a}}_t$ the final action vector $\hat{\mathbf{a}}_t$ is generated.
This design allows the model to maintain a fixed-size output from the DT while dynamically adapting to any number of nodes (and hence actions). It effectively combines the high-level context learned by the transformer with the node-specific state information captured by the GNN, enabling robust, scalable decision-making even as the graph structure changes.


\subsection{Training: Action Masking and Loss Function }

The proposed GNN-DT model is trained via supervised learning using an offline trajectory dataset~\cite{chen2021decision}, similarly to offline RL. Specifically, the GPT-2 model is initialized with its default pre-trained weights, which are subsequently fine-tuned end-to-end for the EV charging optimization task.
In GNN-DT, the learning of infeasible actions, such as charging an unavailable EV, is avoided through action masking. At each time step $t$, a mask vector $\mathbf{m}_t$, which has the same dimension as $\mathbf{a}_t$, is generated with zeros marking invalid actions and ones marking valid actions. For example, an action is invalid when the $a_{i,t} \neq 0$ and no EV is connected at charger $i$.
The mean squared error between the predicted actions $\widehat{\mathbf{a}}_t$ and ground-truth actions $\mathbf{a}_t$ from expert or offline trajectories is employed as the loss function. For a window of length $K$ ending at time $t$, training loss is defined as:
\begin{equation}
  \mathcal{L}
  = \frac{1}{K}
    \sum\nolimits_{\tau=t-K}^{t}
    \bigl\lVert (\widehat{\mathbf{a}}_{\tau}-\mathbf{a}_\tau)\circ\mathbf{m}_\tau\bigr\rVert^2.
\end{equation}
By incorporating the mask into the loss calculation (elementwise multiplication), a focus solely on valid actions is enforced, thereby preserving meaningful gradient updates.




\section{Experimental Setup}
The dataset generation and the evaluation experiments are conducted using the EV2Gym simulator~\cite{10803908}, which leverages real-world data distributions, including EV arrivals, EV specifications, electricity prices, etc. This setup ensures a realistic environment where the state and action spaces accurately reflect real charging stations’ operational complexity. A scenario with 25 chargers is chosen, allowing up to 25 EVs to be connected simultaneously. In this configuration, the action vector has up to 25 variables (one per EV), while the state vector contains around 150 variables describing EV statuses, charger conditions, power transformer constraints, and broader environmental factors.
Consequently, the resulting optimization problem is in the moderate-to-large scale range, reflecting the key complexities of real-world EV charging. Each training procedure is repeated 10 times with distinct random seeds to ensure statistically robust findings. All reported rewards represent the average performance over 50 evaluation scenarios, each featuring different configurations (electricity prices, EV behavior, power limits, etc.). 
Training was carried out on an NVIDIA A10 GPU paired with 11 CPU cores and 80 GB of RAM, using the AdamW optimizer and a LambdaLR scheduler. Baseline RL agents converged in 2–5 hours, while the proposed GNN-DT required up to 10 hours of training. Default hyperparameters were used for all baseline RL methods; Table \ref{tab:hyperparam_table} lists the full set of hyperparameters employed to train the DTs.

\begin{table}[t]
\centering
\caption{Algorithm hyperparameters for small- and large-scale settings.}
\label{tab:hyperparam_table}
\begin{tabular}{lcc}
\toprule
\textbf{Hyperparameter} & \textbf{Small scale} & \textbf{Large scale} \\
\midrule
Batch size                 & 128      & 64    \\
Learning rate              & $10^{-4}$& $10^{-4}$ \\
Weight decay               & $10^{-4}$& $10^{-4}$ \\
Steps per iteration        & 1000     & 3000  \\
Decoder layers             & 3        & 3     \\
Attention heads            & 4        & 4     \\
Embedding dimension        & 128      & 256   \\
GNN embedder feat.\ dim.   & 16       & 16    \\
GNN hidden dimension       & 32       & 64    \\
GCN layers                 & 3        & 3     \\
Epochs                     & 250      & 400   \\
CPU memory (GB)            & 8        & 40    \\
Time limit (h)             & 10       & 46    \\
\bottomrule
\end{tabular}%
\end{table}

\begin{figure*}[!t]     
    \centering
    \includegraphics[width=1\textwidth]{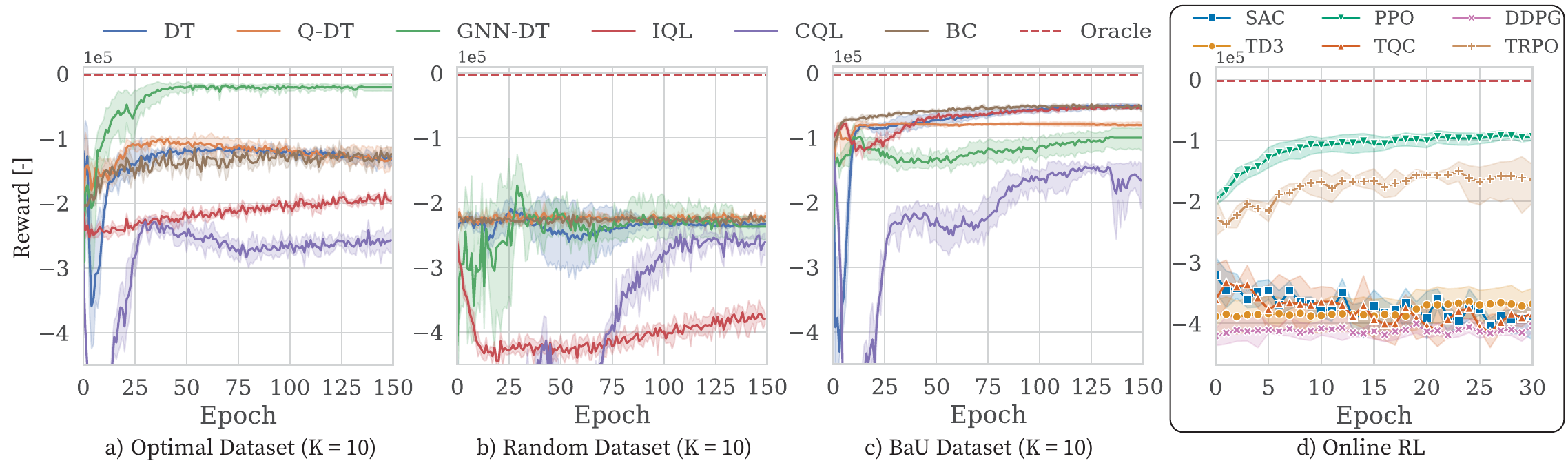}
        \caption{Training performance comparison for online and offline RL algorithms.
        }
        \label{fig:perf_comp}
\end{figure*}

\subsection{Dataset Generation}
\label{subsec:dataset_generation}

Offline RL algorithms, including DTs, can learn policies from trajectories without needing online interaction with the environment. Consequently, the quality of the gathered training trajectories has a substantial impact on the learning process. In this work, three distinct strategies were used to generate trajectories:
\begin{itemize}
    \item \textbf{Random Actions}: Uniformly sampled actions in the range $[-1, 1]$ were applied to the simulator.
    \item \textbf{Business-as-Usual (BaU)}: A Round Robin charging policy commonly employed by CPOs, which sequentially allocates charging power among EVs to balance fairness and efficiency.
    \item \textbf{Optimal Policies}: Optimal solutions derived from solving offline the mathematical problem described in Section~\ref{sec:mip} for randomly generated scenarios.
\end{itemize}
Each trajectory consists of 300 state-action-reward-action mask tuples, with each timestep representing a 15-minute interval, resulting in a total of three simulated days. This combination of random, typical, and expert data provides a comprehensive basis for evaluating how GNN-DT learns from diverse offline trajectories.

\section{Experiments}
\label{sec:experiments}

In this section, a comprehensive set of experiments is presented to evaluate the proposed method’s performance, both during training and under varied test conditions. Different dataset types and sample sizes are examined to determine their impact on learning efficiency and convergence.

\subsection{Training Performance}

Fig.~\ref{fig:perf_comp} compares the proposed GNN-DT against multiple baselines, including the {classic DT}~\cite{chen2021decision}
and {Q-DT}~\cite{qdt}
, which both rely on flattened state representations due to their inability to directly process graph-structured data. In these baseline methods, empty chargers and unavailable actions are replaced by zeros, so the action vector is always the same size. Several well-known online RL algorithms from the Stable-Baselines-3~\cite{sb3} framework are evaluated, such as SAC, DDPG, Twin Delayed DDPG (TD3), Trust Region Policy Optimization (TRPO), Proximal Policy Optimization (PPO), and Truncated Quantile Critics (TQC). Also, offline RL algorithms from D3RLPY~\cite{d3rlpy}, namely Implicit Q-Learning (IQL), Conservative Q-Learning (CQL), and Behavioral Cloning (BC), are also included.
The offline RL algorithms (IQL, CQL, BC, DT, Q-DT, and GNN-DT) are trained on three datasets (\textit{Optimal}, \textit{Random}, and \textit{BaU}), each comprising 10.000 trajectories. 
A red dotted line marks the optimal reward, which represents the experimental maximum achievable reward obtained by solving the deterministic MIP knowing the future (\textit{Oracle}) defined in Eq.~\ref{eq:mo4}. This oracle reward serves as an upper bound and helps contextualize the relative performance of each method. It is important to note that the EV smart charging problem requires real-time (1-5 min intervals) optimization solutions for large-scale, highly stochastic scenarios, where metaheuristic algorithms, stochastic optimization, and model predictive control methods fail due to computational constraints. By contrast, once trained (over 2–24 hours), RL agents can deliver real-time charging schedules on an ordinary computer in milliseconds.

\begin{table*}[t]
\centering
\small
\caption{Comparison of maximum episode rewards ($\times10^{5}$) for baselines and GNN-DT across datasets and context lengths ($K$). \textbf{Bold} indicates the highest value within each dataset and $K$ category.}
\label{tab:rewards_ov}
\begin{tblr}{
  row{even} = {c},
  row{3} = {c},
  row{5} = {c},
  row{7} = {c},
  row{9} = {c},
  row{11} = {c},
  cell{1}{1} = {r=2}{c},
  cell{1}{2} = {r=2}{},
  cell{1}{3} = {r=11}{},
  cell{1}{4} = {c=3}{c},
  cell{1}{7} = {c=3}{c},
  cell{3}{2} = {r=3}{},
  cell{6}{2} = {r=3}{},
  cell{9}{2} = {r=3}{},
  hline{1,12} = {-}{0.08em},
  hline{2} = {4-9}{0.03em},
  hline{3} = {1-2,4-9}{0.03em},
  hline{6,9} = {1-2,4-9}{dashed},
}
Dataset & {{Avg. Training}\\{Dataset Reward}} &  & K=2 &  &  & K=10 &  & \\
 &  &  & DT & Q-DT & GNN-DT & DT & Q-DT & GNN-DT \\
Random $100$ & $-2.37$~\textcolor[rgb]{0.216,0.212,0.216}{±$0.39$} &  & $-1.91$ & $-1.97$ & $\mathbf{-0.82}$ & $-2.12$ & $-2.09$ & $-1.16$\\
Random $1000$ &  &  & $-1.93$ & $-2.04$ &$ -0.86$ & $-2.11$ & $-2.01$ & $-1.18$\\
Random $10000$ &  &  & $-1.76$ & $-2.04$ & $-1.25$ & $-1.81$ & $-1.98$ & $\mathbf{-0.98}$\\
BaU $100$ & $-0.67$~\textcolor[rgb]{0.216,0.212,0.216}{±$0.07$} &  & $-0.79$ & $-0.74$ & $\mathbf{-0.59}$ & $-0.79$ & $-0.72$ & $-0.56$\\
BaU $1000$ &  &  & $-0.71$ & $-0.66$ & $-0.65$ & $-0.64$ & $-0.71$ &$ -0.57$\\
BaU $10000$ &  &  & $-0.69$ & $-0.66$ & $-0.66$ & $\mathbf{-0.44}$ & $-0.74$ & $-0.53$\\
Optimal $100$ & $-0.01$~\textcolor[rgb]{0.216,0.212,0.216}{±$0.01$} &  & $-0.67$ & $-0.91 $& $-0.15$ & $-1.12$ & $-0.90$ & $-0.14$\\
Optimal $1000$ &  &  & $-0.63$ & $-0.67 $& $-0.10$ & $-0.87$ & $-0.86$ & $-0.09$\\
Optimal $10000$ &  &  & $-0.63$ & $-0.80 $& $\mathbf{-0.04}$ & $-0.72$ & $-0.90$ & $\mathbf{-0.07}$
\end{tblr}
\end{table*}

In Figs.~\ref{fig:perf_comp}.a-c, the DT-based approaches use a context length $K=10$. As expected, the \emph{Optimal} dataset provides the highest-quality information, enabling GNN-DT to converge rapidly toward near-oracle performance, while classic DT, Q-DT, and the other offline RL algorithms lag far behind, showcasing GNN-DT's improved sample efficiency. With the \emph{Random} dataset, the limited quality of data leads all methods to plateau at lower reward values, although GNN-DT still surpasses the other baselines. An intriguing behavior is observed with the \emph{BaU} dataset, where classic DT, BC, and IQL converge at rewards exceeding those of GNN-DT. In contrast, the online RL algorithms displayed in Fig.~\ref{fig:perf_comp}.d struggle to achieve comparable improvements, suggesting that pure online exploration is insufficient for solving this complex EV charging optimization problem with sparse rewards. 
In the rest of this section, we omit the online and offline RL baselines, as their performance is substantially inferior to that of DT, Q-DT, and GNN-DT.

\subsection{Dataset Impact}
In Table~\ref{tab:rewards_ov}, the maximum episode reward is compared for small, medium, and large datasets (100, 1.000, and 10.000 trajectories), under two different context lengths ($K=2$ and $K=10$). The left side of Table~\ref{tab:rewards_ov} reports the dataset type, the number of trajectories, and the average reward in each dataset. All baselines achieve performance above the \emph{Random} dataset’s average reward. However, only GNN-DT consistently approaches the \emph{Optimal} dataset’s performance, reaching as close as $-0.04\times10^{5}$ compared to the $-0.01\times10^{5}$ optimal reward. This advantage becomes especially evident at the largest dataset size (10.000 trajectories), highlighting the benefits of the graph-based embedding layer. Overall, GNN-DT outperforms the baselines across all datasets and both context lengths, with the single exception of the \emph{BaU} dataset at $K=10$. Interestingly, a larger context window does not always translate into higher rewards, potentially due to the problem setting. Similarly, the dataset size appears to have minimal impact on Q-DT, whereas DT and GNN-DT generally improve with more trajectories. These findings underscore that both the quality and quantity of offline data, coupled with the GNN-DT architecture, are key to achieving superior performance.

\subsection{Enhancing Training Datasets}

The previous section highlighted that the quality of trajectories in the training dataset is the most influential factor for achieving high performance. In this section, we explore whether creating new datasets by mixing existing ones can further improve performance. The \emph{Optimal} and \emph{Random} datasets are combined in different proportions, as summarized in Table~\ref{tab:mixed_opt}. A noteworthy result is that supplementing the \emph{Optimal} dataset with ``less useful'' (\emph{Random}) trajectories consistently boosts performance. In particular, GNN-DT with $K=10$, trained on a mix of 250 \emph{Optimal} and 750 \emph{Random} trajectories, achieves near-oracle results, deviating by only $-0.001\times10^{5}$ from the optimal reward.
A similar trend emerges when blending \emph{BaU} and \emph{Random} datasets shown in Table~\ref{tab:mixed_bau}. While the BaU dataset alone performs worse than the Optimal dataset, mixing it with Random data still yields improvements, with the 75\% BaU and 25\% Random combination showing the best results.
Overall, these findings indicate that carefully integrating high- and lower-quality data can enhance policy learning beyond what purely \emph{Optimal} or purely \emph{Random} datasets can provide.

\begin{table*}[!t]
\centering
\caption{Maximum reward of GNN-DT trained on merged \textit{Optimal} and \textit{Random} datasets for $K=2$ and $K=10$. Performance improves despite lower average training rewards, highlighting the importance of dataset diversity.
Highest rewards per $K$ are highlighted with \textbf{bold}.}
\label{tab:mixed_opt}
\begin{tabular}{ccc|cc} 
\toprule
\multirow{2}{*}{Dataset} & \multirow{2}{*}{\begin{tabular}[c]{@{}c@{}}Total \\Traj.\end{tabular}} & \multicolumn{1}{c}{\multirow{2}{*}{\begin{tabular}[c]{@{}c@{}}Avg. Dataset\\Reward\end{tabular}}} & \multicolumn{2}{c}{GNN-DT Reward ($\times10^5$)}      \\ 
\cline{4-5}
                         &                                                                        & \multicolumn{1}{c}{}                                                                              & K=2               & K=10               \\ 
\hline
Random (Rnd.) 100\%      & $1000$                                                                 & $-2.37$ ±$0.39$                                                                                   & $-0.863$          & $-1.187$           \\
Opt. 25\% + Rnd. 75\%    & $1000$                                                                 & $-1.78$~±$1.07$                                                                                   & $-0.045$          & $\mathbf{-0.020}$  \\
Opt. 50\% + Rnd. 50\%    & $1000$                                                                 & $-1.18$~±$1.19$                                                                                   & $\mathbf{-0.021}$ & $-0.040$           \\
Opt. 75\% + Rnd. 25\%    & $1000$                                                                 & $-0.60~$±$1.03$                                                                                   & $-0.073$          & $-0.057$           \\
Optimal (Opt.) 100\%     & $1000$                                                                 & $-0.01$~±$0.01$                                                                                   & $-0.108$          & $-0.099$           \\
\bottomrule
\end{tabular}
\end{table*}

\begin{table*}[!t]
\centering
\small
\caption{Maximum reward of GNN-DT trained on merged BaU-Random datasets for $K=2$ and $K=10$.The bold indicates the training dataset with the highest evaluation reward.}
\label{tab:mixed_bau}
\begin{tabular}{ccc|cc} 
\toprule
\multirow{2}{*}{Dataset} & \multirow{2}{*}{\begin{tabular}[c]{@{}c@{}}Total \\Traj.\end{tabular}} & \multicolumn{1}{c}{\multirow{2}{*}{\begin{tabular}[c]{@{}c@{}}Avg. Dataset\\Reward\end{tabular}}} & \multicolumn{2}{c}{GNN-DT Reward ($\times10^5$)}      \\ 
\cline{4-5}
                         &                                                                        & \multicolumn{1}{c}{}                                                                              & K=2               & k=10               \\ 
\hline
Random (Rnd.) 100\%      & $1000$                                                                 & $-2.37$ ±$0.39 $                                                                                  & $-0.863$          & $-1.187$           \\
BaU~25\% + Rnd. 75\%     & $1000$                                                                 & $-1.93$ ±$0.80$                                                                                   & $-0.578$          & $-0.461$           \\
BaU~50\% + Rnd. 50\%     & $1000$                                                                 & $-1.51$ ±$0.87$                                                                                   & $-0.665$          & $\mathbf{-0.447}$  \\
BaU~75\% + Rnd. 25\%     & $1000$                                                                 & $-1.09$ ±$0.76$                                                                                   & $\mathbf{-0.421}$ & $-0.471$           \\
BaU~100\%                & $1000$                                                                 & $-0.01$~$±0.01$                                                                                   & $-0.654$          & $-0.572$           \\
\bottomrule
\end{tabular}
\end{table*}

\begin{figure*}[!t]
\centering
\includegraphics[width=0.5\linewidth]{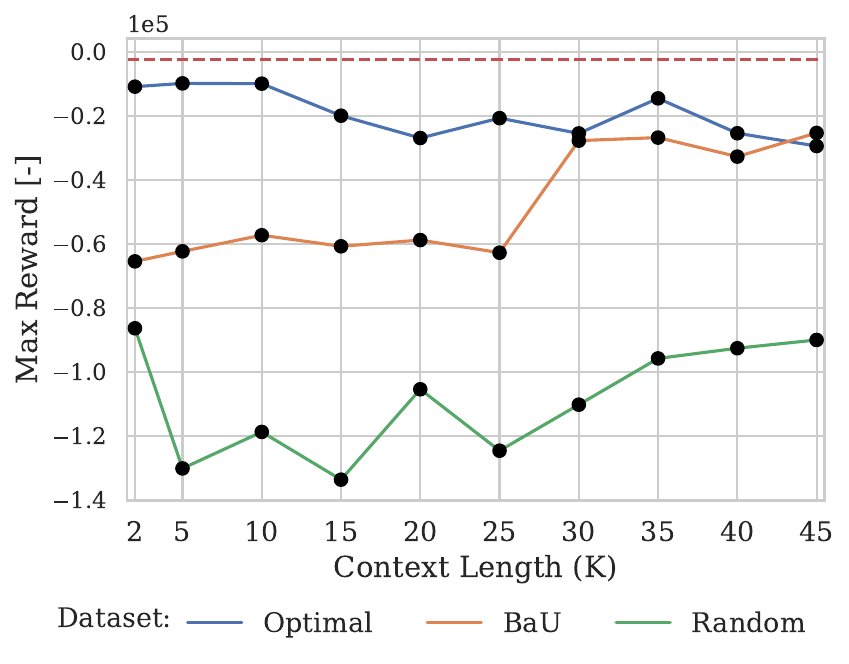}
\caption{
GNN-DT performance for larger context lengths (K).}
\label{fig:K_comp}
\end{figure*}

\begin{table*}[h]
\centering
\small
\caption{Average reward trained over 5 runs with different seeds for \textit{Optimal} and \textit{Mixed} datasets.}
\label{tab:ablation}
\begin{tblr}{
  cells = {c},
  hline{1,8} = {-}{0.08em},
  hline{2} = {-}{0.05em},
}
DT & State GNN & Action GNN & Res. Con. & Action Mask & Optimal ($\times 10^5$) & Mixed ($\times 10^5$) \\
\cmark  & \xmark          & \xmark           & \xmark          & \xmark            & $-0.69 \pm 0.03$        & $-0.95 \pm 0.39$       \\
\cmark  & GCN       & \xmark           & \xmark          & \xmark            & $-0.71 \pm 0.02$        & $-0.77 \pm 0.17$       \\
\cmark  & GCN       & \xmark           & \cmark         & \xmark            & $-0.18 \pm 0.03$        & $-0.16 \pm 0.07$       \\
\cmark  & GCN       & GCN        & \cmark         & \xmark           & $-0.11 \pm 0.03$        & $-0.12 \pm 0.04$       \\
\cmark  & GAT       & GAT        & \cmark         & \cmark           & $-0.14 \pm 0.07$        & $-0.15 \pm 0.06$       \\
\cmark  & GCN       & GCN        & \cmark         & \cmark           & $\mathbf{-0.09 \pm 0.02}$ & $\mathbf{-0.10 \pm 0.04}$\\
\end{tblr}
\end{table*}

\subsection{Impact of larger context lengths (K)}
Fig.~\ref{fig:K_comp} demonstrates that the context length $K$ plays a key role in the performance of GNN-DT, with diminishing returns beyond a certain point. For high-quality datasets like \textit{Optimal}, moderate context lengths ($K=5$ to $K=10$) yield the best results, while larger $K$ values do not improve performance significantly. For suboptimal datasets like \textit{BaU} and \textit{Random}, the performance is lower overall, and longer context lengths seem to offer meaningful improvements, particularly when using the \textit{BaU} dataset. Thus, selecting an appropriate context length is crucial for achieving better performance, while the quality of the dataset remains the most influential factor.

\subsection{Component Ablation Study}
To better understand the contribution of each architectural component, we conduct an ablation study that systematically removes or replaces elements of our model and then trains the model on the \textit{Optimal} and \textit{Mixed} (Opt.25\% +Rand.75\%). The results in Table~\ref{tab:ablation} reveal that neither a plain DT nor a DT augmented solely with a state-GNN submodule achieves competitive performance. 
Notably, adding the residual connection atop the state-GNN leads to a significant improvement, from $-0.77\times10^5$ to $-0.16\times10^5$ on the \textit{Mixed} dataset, demonstrating its importance for effective credit assignment over dynamic inputs. Removing action masking or replacing the GCN module with a Graph Attention Network (GAT)~\cite{veličković2018graphattentionnetworks} similarly degrades performance, indicating that each component provides distinct and complementary benefits. Ultimately, only the full GNN-DT architecture achieves strong performance across both the \textit{Mixed} and \textit{Optimal} datasets.

\subsection{Average Results of EV Charging.}
\label{sec:res}

Table~\ref{tab:ev_exps_results} shows a comparison of key EV charging metrics for the 25-station problem after 100 evaluations, including heuristic algorithms, Charge As Fast as Possible (CAFAP)  and BaU, and DT variants with the optimal solution, which assumes future knowledge.

The performance of the proposed algorithms was assessed using several evaluation metrics. 
For example, user satisfaction [\%] captures the extent to which the state of charge at departure (\(e_{j,t^d}\)) of each electric vehicle \(j \in \mathcal{J}\) meets its target \(e_j^*\), thus defined as:  
\begin{equation}
\text{User Satisfaction [\%]}  = \frac{1}{|\mathcal{J}|} \sum_{j \in \mathcal{J}} \left( \frac{e_{j,t^d}}{e_j^*} \right) \cdot 100 \%.
\end{equation}
Energy charged [kWh] was measured as the total amount of energy delivered to the vehicles during the charging sessions, while energy discharged [kWh] was quantified as the energy returned from vehicles to the grid. Power violations [kW] were tracked to identify instances in which operational limits were exceeded, ensuring system feasibility. Finally, the overall charging cost [€] was evaluated by accounting for the time-varying electricity prices during charging and discharging periods, thus reflecting the economic performance of the strategy.

GNN-DT shows remarkable performance, achieving a close approximation to the optimal solution, particularly in user satisfaction (99.3\% ± 0.03\%) and power violation (21.7 ± 22.8 kW). It outperforms both BaU and DT variants in terms of energy discharged, power violation, and costs. Notably, GNN-DT performs well even compared to Q-DT, while maintaining competitive execution time, albeit slightly slower than the simpler models. The results underscore the effectiveness of GNN-DT in managing complex EV charging tasks, demonstrating its potential for real-world applications where future knowledge is not available.

\begin{table*}[!h]
\centering
\small
\caption{Comparison of key EV charging metrics for the 25-station problem after 100 evaluations, for heuristic algorithms (CAFAP \& BaU) and DT variants with the optimal solution, which assumes future knowledge. }
\label{tab:ev_exps_results}
\begin{tblr}{
  cells = {c,t},
  vline{2} = {2-7}{0.05em},
  hline{1,8} = {-}{0.08em},
  hline{2} = {-}{0.05em},
}
Algorithm & {Energy\\~Charged\\~[MWh]} & {Energy\\~Discharged\\~[MWh]} & {User\\~Satisfaction\\~[\%]} & {Power\\~Violation\\~[kW]} & {Costs\\~[€]} & {Reward\\~[-$10^5$]} & {Exec. Time\\~[sec/step]}\\
CAFAP & $1.3$ ±$0.2$ & $0.00$ ±$0.00$ & $100.0$ ±$0.0$ & $1289.2$ ±$261.8$ & $-277$ ±$165$ & $-1.974$ ±$0.283$ & $0.001$\\
BaU & $1.3$ ±$0.2$ & $0.00$ ±$0.00$ & $99.9$ ±$0.2$ & $10.5$ ±$9.4$ & $-255$ ±$156$ & $-0.679$ ±$0.067$ & $0.001$\\
DT & $0.9$ ±$0.1$ & $0.03$ ±$0.01$ & $94.4$ ±$1.6$ & $58.7$ ±$28.3$ & $-173$ ±$104$ & $-0.462$ ±$0.093$ & $0.006$\\
Q-DT & $1.0$ ±$0.1$ & $0.00$ ±$0.00$ & $93.6$ ±$2.1$ & $20.1$ ±$21.4$ & $-187$ ±$113$ & $-0.665$ ±$0.135$ & $0.010$\\
\textbf{GNN-DT} (Ours) & $0.9$ ±$0.1$ & $0.19$ ±$0.03$ & $99.3$ ±$0.2$ & $21.7$ ±$22.8$ & $-142$ ±$89$ & $-0.027$ ±$0.023$ & $0.023$\\
Optimal (Offline) & $1.9$ ±$0.2$ & $1.08$ ±$0.19$ & $99.1$ ±$0.2$ & $2.0$ ±$4.6$ & $-119$ ±$84$ & $-0.020$ ±$0.015$ & -
\end{tblr}
\end{table*}

\begin{figure*}[!t]     
\centering
     \subfloat[EVs' battery level progression over time.]{
         \centering
         \includegraphics[width=0.48\linewidth]{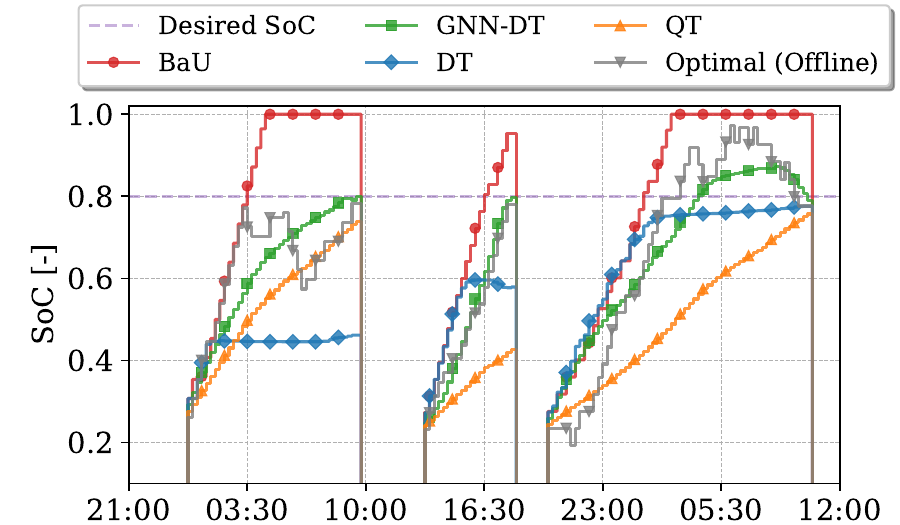}
         \label{fig:comp_a}
         }
     \subfloat[Charger Actions]{
         \centering
         \includegraphics[width=0.48\linewidth]{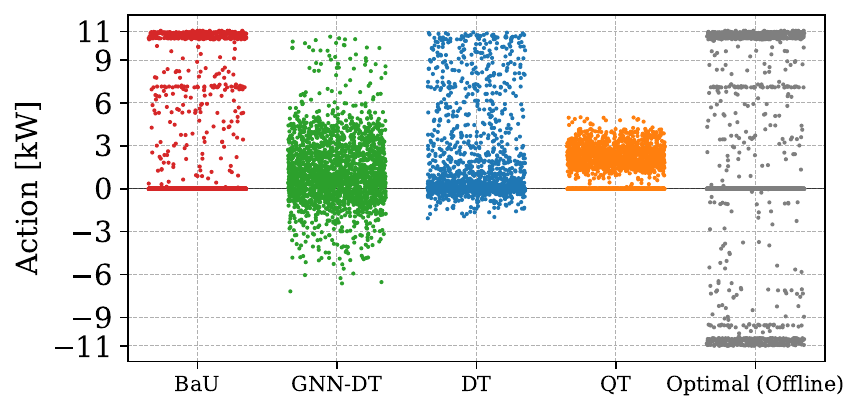}
         \label{fig:comp_b}
         }
         
        \caption{Comparison of smart charging algorithms for a single simulation day.
        }
        \label{fig:physics}
\end{figure*}

\subsection{Illustrative Example of EV Charging}

With the models trained, we proceed to compare the behavior of the best baseline models trained (DT, Q-DT, GNN-DT) against the heuristic BaU algorithm in an EV charging scenario. Fig.~\ref{fig:comp_a} presents the SoC progress for three EVs connected one after the other to a single charger throughout the simulation, while Fig.~\ref{fig:comp_b} shows the actions of all chargers taken by each algorithm.
At the beginning of the simulation, EVs arrive at the charging station with unknown initial SoCs. Upon connection, they communicate their departure times and desired SoC levels to the CPO. Leveraging this information, along with real-time electricity price signals and power constraints, each algorithm determines optimal charging and discharging actions.

In Fig.~\ref{fig:comp_a}, the heuristic {BaU} algorithm consistently overcharges the EVs, often exceeding the desired SoC levels. In contrast, both DT and Q-DT fail to satisfy the desired SoC. Conversely, GNN-DT successfully achieves the desired SoC for all EVs, closely mirroring the behavior of the optimal algorithm. This demonstrates GNN-DT's ability to precisely control charging based on dynamic state information. Fig.~\ref{fig:comp_b} provides further insights into the actions taken by each algorithm. The optimal solution primarily employs maximum charging or discharging power, since it knows the future. In comparison, GNN-DT exhibits a more refined approach, modulating charging power within a range of -6 to 11 kW.  
Baseline DT and Q-DT display a narrower range of actions, limiting their ability to optimize the charging schedules and adapt to varying conditions.
These results underscore the superior capability of GNN-DT in managing the complexities of EV charging dynamics.

\subsection{Generalization and Scalability Analysis}
\label{sec:lim}


\begin{figure*}[!h]
  \centering
  \subfloat[Arrival time]{%
    \includegraphics[width=0.32\textwidth]{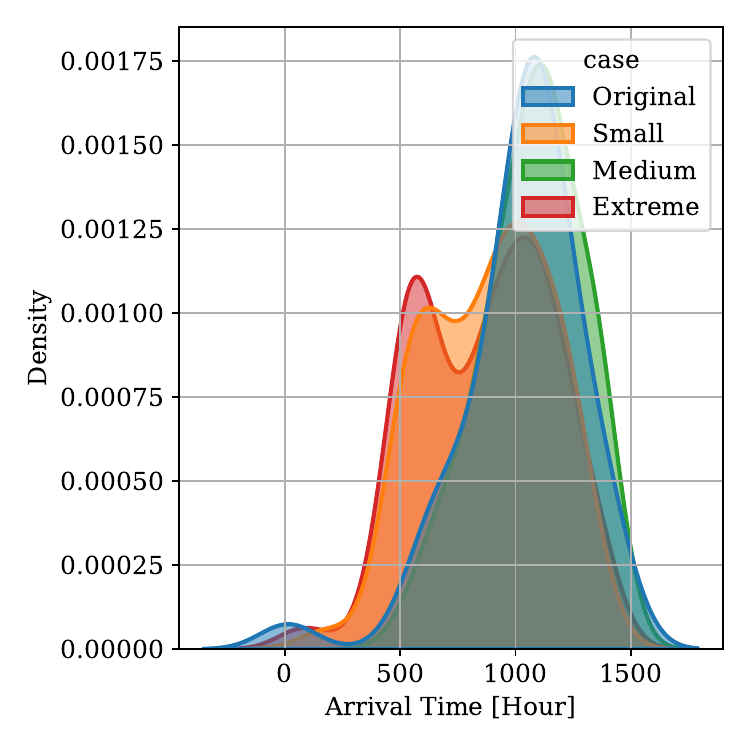}%
    \label{fig:arrival_time}%
  }\hfill
  \subfloat[Departure time]{%
    \includegraphics[width=0.32\textwidth]{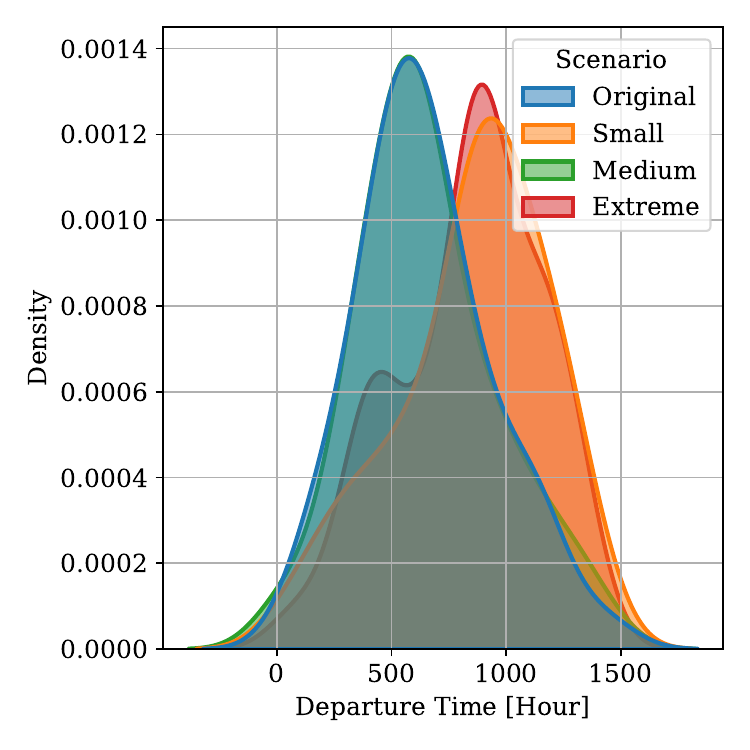}%
    \label{fig:departure_time}%
  }\hfill
    \subfloat[Time of stay]{%
    \includegraphics[width=0.32\textwidth]{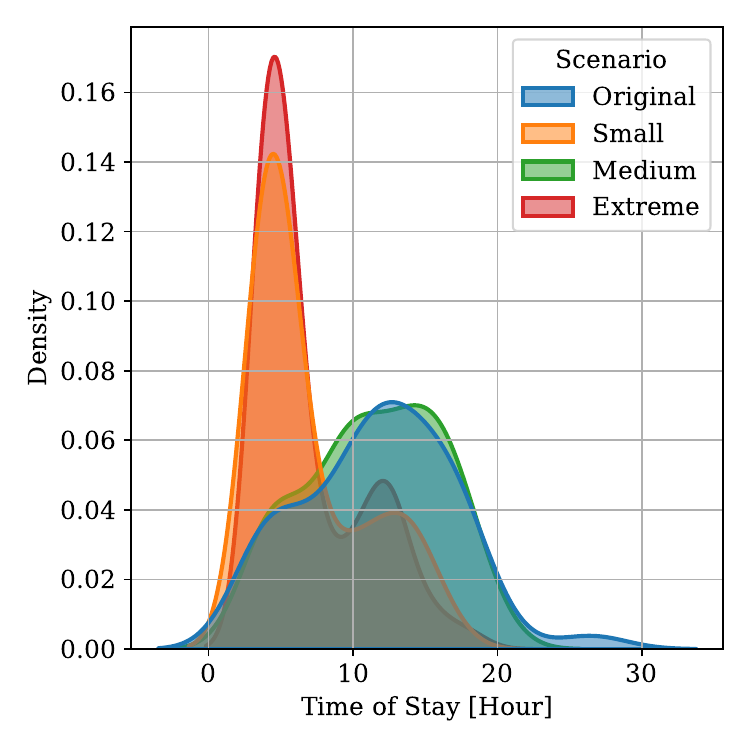}%
    \label{fig:time_of_stay}%
  }
  \\[1ex] 
  \subfloat[SoC at arrival]{%
    \includegraphics[width=0.32\textwidth]{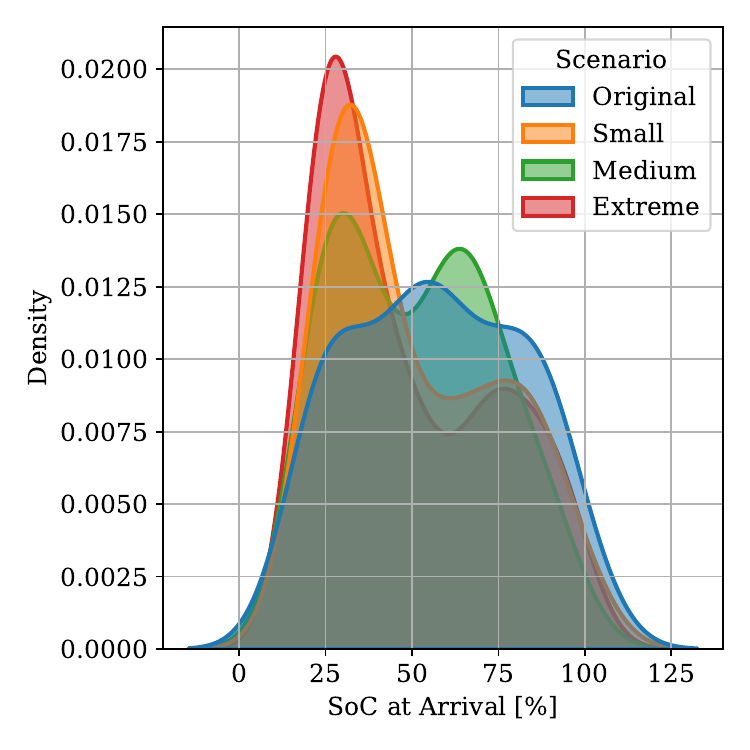}%
    \label{fig:soc_at_arrival}%
  }\hfill
  \subfloat[Power limit]{%
    \includegraphics[width=0.65\textwidth]{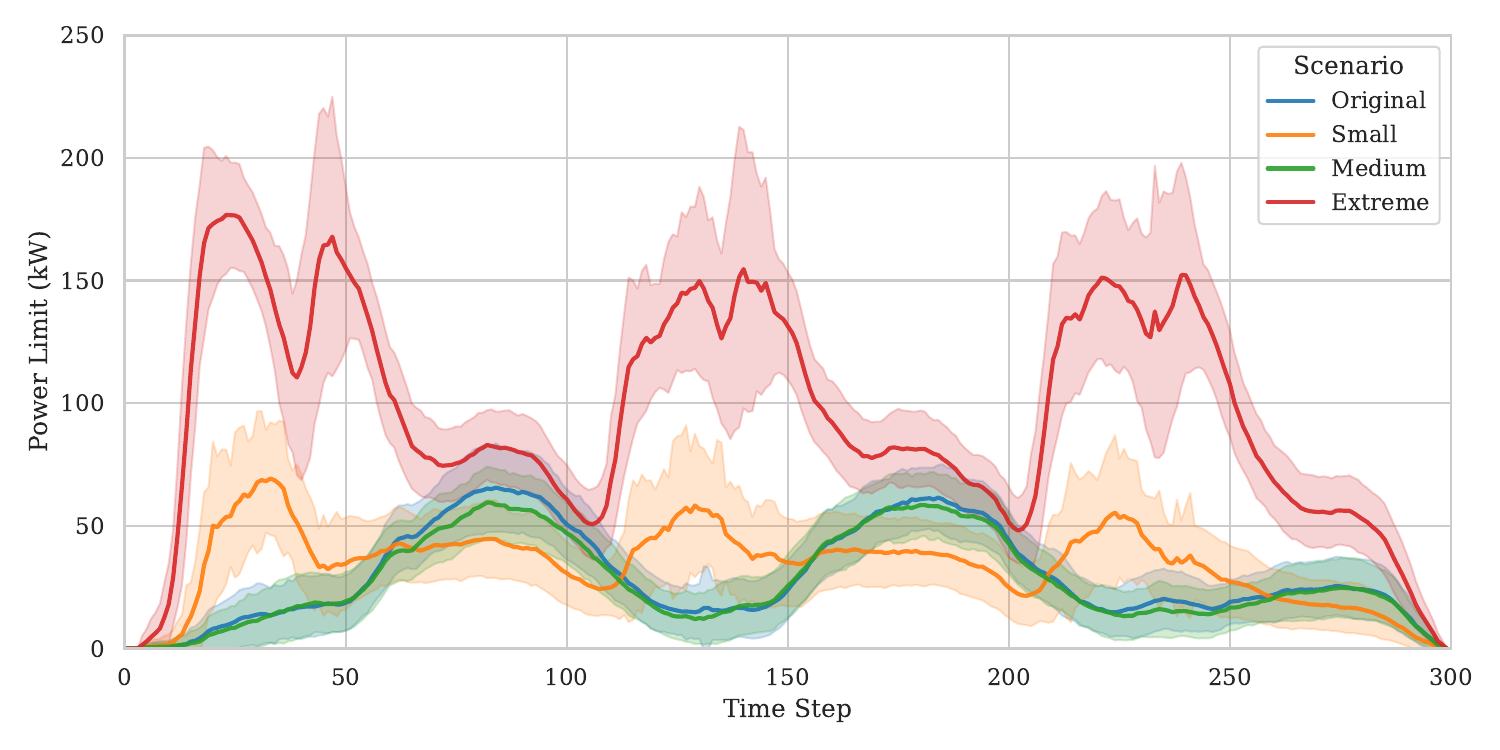}%
    \label{fig:power_limit}%
  }
  \caption{Overview of the five key state transition variables across different scenarios: (a) arrival time, (b) departure time, (c) power limit, (d) state of charge upon arrival, and (e) time of stay.}
  \label{fig:gen_prob}
\end{figure*}

\begin{figure*}[!h]
  \centering
  \subfloat[Unseen state transition probabilities.]{
    \includegraphics[width=0.48\linewidth]{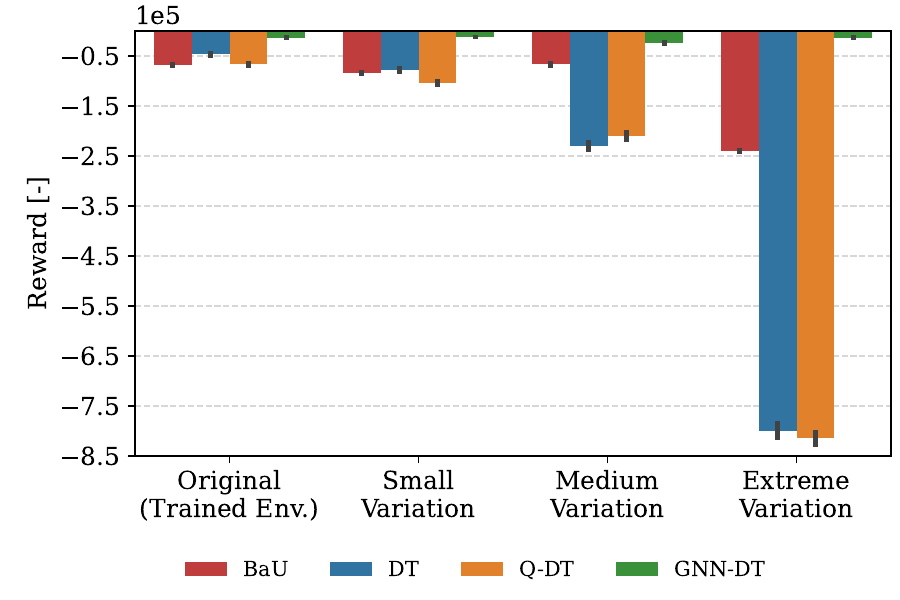}
    \label{fig:gen_a}
  }
  \subfloat[Evaluating on different size problems.]{
    \includegraphics[width=0.48\linewidth]{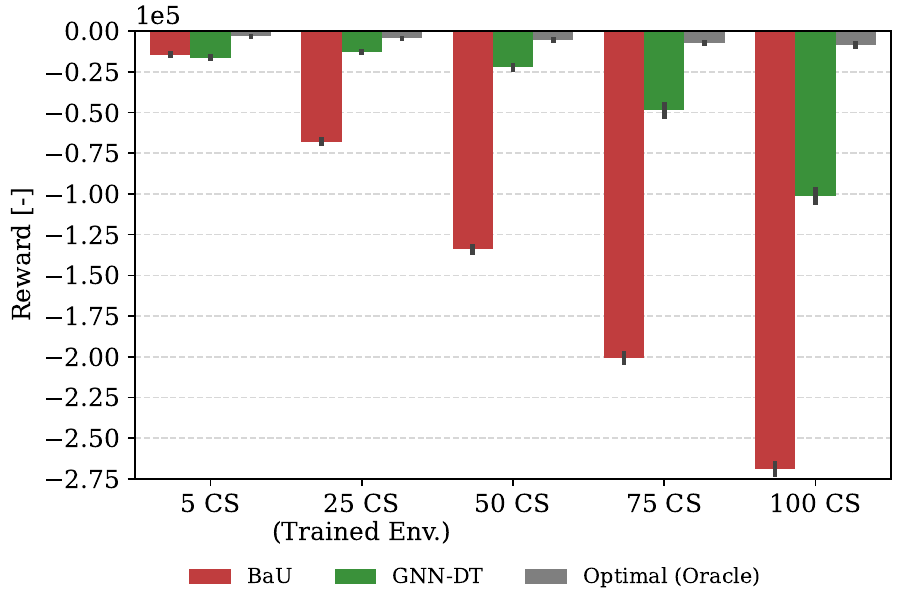}
    \label{fig:gen_b}
  }
  \caption{Generalization performance of the proposed model, depicting the average rewards achieved across 100 randomly generated scenarios in previously unseen environments.}
  \label{fig:generalization}
\end{figure*}

Evaluating the generalization of RL models across varying state transition probabilities is crucial for ensuring consistent performance under diverse conditions~\cite{wang2020statistical}. 
To evaluate the generalization capabilities of GNN-DT, three additional environments with different state transition probabilities are designed.
The key environment variables that directly impact the state transition dynamics are visualized in Figure~\ref{fig:gen_prob}. In detail, Figure~\ref{fig:gen_prob}a–d presents the probability distributions of EV arrival time, departure time, duration of stay, and state of SoC at arrival across four scenarios: the original training environment and environments with small, medium, and extreme variations. These plots help quantify the extent of variation in each case. Additionally, Figure~\ref{fig:gen_prob}e illustrates the temporal distribution of the power limit in each scenario, providing further insight into the differences in environment configuration.

In Fig.~\ref{fig:gen_a}, the generalization capabilities of GNN-DT and other baselines are assessed in environments with small, medium, and extreme variations in state transition probabilities.
While the baseline methods experience significant performance drops as the evaluation environment deviates from the training setting, GNN-DT maintains strong performance across all scenarios. This highlights the critical role of GNN-based embeddings in improving model robustness and generalization.
A key advantage of the GNN-DT architecture, not present in classic DTs, is its invariance to problem size, i.e., the same RL agent can be applied to both smaller and larger-scale environments. Fig.~\ref{fig:gen_b} illustrates the scalability and generalization performance of GNN-DT compared to the BaU algorithm and Optimal policy.
GNN-DT, trained on a 25-charger setup, was tested on 5, 50, 75, and 100-charger environments. While its performance predictably declines at larger scales, since it wasn’t trained on them, it still outperforms the BaU heuristic, demonstrating robustness to problem-size variation. Training GNN-DT on a mix of charger numbers could potentially further improve its adaptability.


The scalability and effectiveness of GNN-DT were tested when trained on a significantly larger optimization problem involving 250 charging stations. In this scenario, the model must handle up to 250 action variables per step and over 1,000 state variables, which include critical information such as power limits and battery levels. The results presented in Table~\ref{tab:250reward} demonstrate that GNN-DT shows promise for addressing more complex optimization tasks. However, the model requires a substantial increase in both the number of training trajectories and memory resources to maintain efficiency, highlighting a well-known limitation of DT-based approaches.
Scaling the problem 10× roughly multiplies GPU memory usage, e.g. storing 3,000 trajectories takes $\approx 2$ GB for 25 chargers versus $\approx 20$ GB for 250. While this can bottleneck large‐scale training, parallelization and mini‐batching mitigate it, and overall compute scales with the transformer’s context length K (see Fig.~\ref{fig:K_comp}), pointing to interesting directions for very large problem graphs as future work.

\begin{table}[!h]
\centering
\caption{Max. reward of GNN-DT in a large-scale EV charging optimization task with 250 chargers.}

\label{tab:250reward}
\begin{tabular}{ccc|c} 
\hline
        & \begin{tabular}[c]{@{}c@{}}{Total}\\Trajectories\end{tabular} & \multicolumn{1}{c}{\begin{tabular}[c]{@{}c@{}}Avg. Dataset\\ Reward\end{tabular}} & \begin{tabular}[c]{@{}c@{}}GNN-DT \\Reward\end{tabular}  \\ 
\hline
Random  & $3000$                                                      & $-22.39$~±$1.49$                                                                 & $-9.34$                                                 \\
BaU     & $3000$                                                      & $-6.67$~±$0.32$                                                                  & $-4.23$                                                 \\
Optimal & $3000$                                                      & $-0.08$~±$0.03$                                                                  & $\mathbf{-0.27}$                                                 \\
\hline
\end{tabular}
\end{table}

\section{Conclusions}

In this work, we introduced a novel DT-based architecture, GNN-DT, which incorporates GNN embedders to significantly enhance sample efficiency and overall performance in the sequential decision-making problem of EV charging. Through extensive evaluation across various datasets, including optimal, random, and BaU, we demonstrated that traditional DTs, online and offline RL algorithms fail to effectively solve real-world problems without specialized embeddings. We further show that both the size and quality of input trajectories critically impact the training process, underscoring the importance of carefully selecting datasets for effective learning. Finally, by leveraging the power of GNN embeddings, GNN-DT improved the model’s ability to generalize in previously unseen environments and handle large, complex action spaces. These contributions demonstrate GNN-DT's potential to address complex dynamic optimization challenges beyond EV charging.

\section*{Acknowledgements}
The study was partially funded by the DriVe2X research and innovation project from the European Commission with grant numbers 101056934. The authors acknowledge the use of computational resources of the DelftBlue supercomputer, provided by Delft High Performance Computing Centre (https://www.tudelft.nl/dhpc). This work used the Dutch national e-infrastructure with the support of the SURF Cooperative using grant no. EINF-5716.

\printcredits
 \bibliographystyle{cas-model2-names}
 \bibliography{ref}

\end{document}